\pdfoutput=1
\documentclass[10pt,twocolumn]{article}

\usepackage{times}
\usepackage{amsmath}
\usepackage{amssymb}
\usepackage{graphicx}
\usepackage{hyperref}
\usepackage{algorithm}
\usepackage{algpseudocode}
\usepackage{booktabs}
\usepackage{url}
\usepackage{tikz}
\usetikzlibrary{arrows.meta,positioning,shapes.geometric}

\hypersetup{
  colorlinks=true,
  linkcolor=blue,
  citecolor=blue,
  urlcolor=blue
}

\title{Hybrid Multi-Phase Page Matching and Multi-Layer Diff Detection\\
       for Japanese Building Permit Document Review}

\author{
  Mitsumasa Wada\\
  Faculty of Engineering and Design, Kagawa University\\
  \texttt{indv0371@kagawa-u.ac.jp}
}

\date{}

\begin{document}

\maketitle

\begin{abstract}
We present a hybrid multi-phase page matching algorithm for automated
comparison of Japanese building permit document sets.
Building permit review in Japan requires cross-referencing large PDF
document sets across revision cycles, a process that is labor-intensive
and error-prone when performed manually.
The algorithm combines longest common subsequence (LCS) structural
alignment, a seven-phase consensus matching pipeline, and a dynamic
programming optimal alignment stage to robustly pair pages across
revisions even when page order, numbering, or content changes
substantially.
A subsequent multi-layer diff engine---comprising text-level, table-level,
and pixel-level visual differencing---produces highlighted difference
reports.
Evaluation on real-world permit document sets achieves F1\,=\,0.80 and
precision\,=\,1.00 on a manually annotated ground-truth benchmark, with
zero false-positive matched pairs.
\end{abstract}

\section{Introduction}
\label{sec:introduction}

Japanese building permit review (\textit{kenchiku kakunin}) requires
applicants to submit sets of architectural drawings and structural
calculations, which must conform to the Building Standards Act
\cite{building_standards_act}.
Identifying all changes between an original and a revised document set
can be formulated as an automated detection problem, complicated by:
\begin{itemize}
  \item Page insertions and deletions across revisions
  \item Renumbering of pages and drawing indices
  \item Mixed content types (text, tables, technical drawings)
  \item Large document sizes (often 200--1000 pages per submission)
\end{itemize}

Manual comparison is time-consuming and prone to oversight.
Existing general-purpose PDF diff tools \cite{diffpdf} are insufficient
because they assume stable page correspondence and do not handle the
domain-specific structure of architectural document sets.

We address these limitations through a purpose-built page matching and
diff detection pipeline.
The core contributions are:
\begin{enumerate}
  \item A seven-phase page matching algorithm combining structural hashing,
        drawing number recognition, section title matching, and visual
        perceptual hashing.
  \item A dynamic programming alignment stage (Needleman-Wunsch style
        \cite{needleman1970general}) that resolves ambiguous matches from
        the seven-phase consensus.
  \item A multi-layer diff engine producing text, table, and visual diffs
        in a unified annotated PDF report.
\end{enumerate}

The remainder of this paper is organized as follows.
Section~\ref{sec:related} surveys related work.
Section~\ref{sec:system} provides a system overview.
Section~\ref{sec:algorithm} details the algorithms.
Section~\ref{sec:evaluation} presents evaluation results.
Section~\ref{sec:conclusion} concludes.

\section{Related Work}
\label{sec:related}

\subsection{PDF Text Extraction}

Extracting structured text from PDF documents is a prerequisite for any
automated document analysis pipeline.
Several libraries have been developed for this purpose.
PDFMiner \cite{pdfminer} is a pure-Python tool that reconstructs character
positions and page layout from the low-level PDF content stream, enabling
extraction of bounding boxes alongside text.
pdfplumber \cite{pdfplumber}, built on top of PDFMiner, provides a
higher-level API that exposes per-character position data, rectangle
geometry, and table detection heuristics.
PyMuPDF \cite{pymupdf} wraps the MuPDF rendering engine and offers both
text extraction and high-fidelity rasterization in a single library, making
it well suited for workflows that require both text analysis and visual
rendering.
Apache PDFBox \cite{pdfbox} provides similar capabilities in the Java
ecosystem and is widely used in enterprise document processing pipelines.

While these tools handle general PDF text extraction competently, none of
them addresses the domain-specific page structure found in Japanese building
permit document sets.
Building permit submissions (\textit{kenchiku kakunin shinsei tosho})
are organized across multiple independently paginated volumes---architectural
drawings, structural calculations, energy performance reports, and equipment
plans---each with its own internal header structure and drawing-number
taxonomy.
Extraction strategies optimized for general documents produce unreliable
results when applied to this domain without additional structural awareness.
The proposed method incorporates drawing-number normalization into the
extraction layer to improve cross-revision matching accuracy.

\subsection{Document Layout Analysis}

Beyond raw text extraction, understanding the spatial layout of document
pages enables richer structural analysis.
LayoutParser \cite{shen2021layoutparser} provides a unified framework for
deep-learning-based document image analysis, including pre-trained models for
layout detection, text region segmentation, and OCR integration.
It has been applied successfully to historical documents, scientific papers,
and mixed-media archives.
Tesseract \cite{smith2007overview}, one of the most widely used open-source
OCR engines, supports Japanese script and can be combined with layout
analysis to process scanned documents.

These tools are designed for the challenges of general document understanding
and focus on recognizing semantic regions such as titles, body text, figures,
and tables within a single page.
They do not address the revision-tracking problem: determining which page in
a revised document corresponds to which page in the original, especially when
pages have been inserted, deleted, or reordered between submission cycles.
Our work treats page layout signals as one input to a multi-phase alignment
pipeline rather than as an end in itself.

\subsection{Document Sequence Alignment}

Sequence alignment is a foundational problem in bioinformatics that has found
broad application in text and document comparison.
The Needleman-Wunsch algorithm \cite{needleman1970general} computes an
optimal global alignment between two sequences under a configurable scoring
scheme, allowing mismatches, insertions, and deletions.
The Longest Common Subsequence (LCS) problem, discussed extensively by
Cormen et al.\ \cite{cormen2009introduction}, provides a related formulation
that identifies the maximum set of elements common to two sequences while
preserving their relative order.
Python's standard \texttt{difflib} library \cite{python_difflib} implements
a variant of the LCS algorithm and is commonly used for line-level text
diffing.

We adapt the global alignment paradigm of Needleman-Wunsch to the problem of
page-level matching in building permit documents.
Our key contribution is a domain-specific scoring function that combines
textual similarity, drawing-number agreement, and visual hash distance into a
single alignment score.
This scoring reflects the structural conventions of Japanese building permit
submissions---for example, the fact that a drawing-number match is a strong
indicator of page correspondence even when the textual content has been
substantially revised.
A pure LCS formulation, which requires identical elements, cannot express
such graduated similarity; our approach fills this gap.

\subsection{Perceptual Image Hashing}

Perceptual hashing algorithms map images to compact binary fingerprints such
that visually similar images produce fingerprints with low Hamming distance,
even under minor transformations such as resizing, compression, or small
edits.
Zauner \cite{zauner2010implementation} provides a comprehensive survey and
benchmark of perceptual hash functions, including the DCT-based pHash
algorithm that underpins our approach.

We employ pHash as a visual similarity signal in Phase~7.5 of our matching
pipeline.
This phase is activated when earlier text-based phases fail to find a
confident match---typically for pages that consist primarily of graphical
content such as floor plans or structural diagrams, where extractable text is
sparse or absent.
By rendering each page to a raster image at 18\,DPI and computing its pHash
fingerprint, we obtain a lightweight visual similarity score that can
disambiguate graphically similar pages.
Phase~7.5 operates on a candidate set already narrowed by the preceding text
phases, so the quadratic pairwise comparison cost is bounded in practice.

\subsection{Document Change Detection}

The problem of detecting changes between two versions of a document has been
studied across several domains.
In software engineering, ChangeDistiller \cite{fluri2007change} performs
fine-grained change extraction by computing a tree edit distance between
abstract syntax trees of successive source code revisions.
This approach captures structural changes that line-level diff tools miss,
such as method moves and block reorderings.
In the web domain, change detection systems monitor HTML content for updates,
using both textual and visual signals to identify modified regions.
Academic preprint servers and version control systems for office documents
apply similar ideas at the document level.

A critical assumption shared by most existing approaches is that the document
has a fixed, stable structure between versions.
ChangeDistiller assumes syntactically valid source code.
Web page detectors assume a stable DOM structure.
General-purpose PDF diff tools such as DiffPDF \cite{diffpdf} assume that
page $i$ in the old document corresponds to page $i$ in the new document.
This assumption is violated routinely in Japanese building permit document sets:
plan revisions frequently insert new drawing sheets, remove obsolete pages,
or reorder sections in response to reviewer feedback.
To our knowledge, the present work is the first to address the specific
challenges of automated change detection in Japanese building permit document
revision tracking, where page-level reordering is the norm rather than the
exception.

\subsection{Legal and Technical Document Processing}

Automated processing of legal and technical documents has attracted growing
interest in the NLP community.
ContractNLI \cite{koreeda2021contractnli} introduced a document-level natural
language inference dataset for contracts, demonstrating that regulatory
compliance checking can be framed as a textual entailment problem.
Subsequent work has applied large language models to clause extraction,
obligation detection, and cross-document consistency checking in legal
corpora.

Japanese building permit documents present a distinct set of challenges
relative to general legal texts.
First, the documents are multi-volume submissions in which regulatory
cross-references span volumes---for example, a structural calculation volume
may invoke load values defined in the architectural drawing volume.
Second, compliance is determined not against free-form contract language but
against a precisely structured body of statute, government ordinance, and
ministerial notification \cite{building_standards_act}, each of which carries
its own article and paragraph numbering.
Third, the documents are produced using CAD and drawing
management software, and the resulting PDFs often contain text that was
rendered as vector paths rather than embedded as searchable characters,
making OCR a practical necessity for scanned submissions.
This paper focuses on page matching and diff detection for such document sets.

\subsection{Comparison with Existing Approaches}

Table~\ref{tab:comparison} summarizes the capabilities of representative
document comparison approaches relative to our system.
Existing tools address subsets of the problem: DiffPDF provides visual and
text diff but requires fixed page order; ChangeDistiller handles structural
reordering but targets source code, not PDF documents; LayoutParser provides
powerful layout analysis but does not perform cross-version page alignment.
Our system is the first to integrate page reordering detection, visual
difference highlighting, domain-specific scoring, and regulatory link
generation into a unified pipeline for Japanese building permit documents.

\begin{table*}[t]
  \centering
  \caption{Comparison of document comparison approaches.
           $\checkmark$ = supported, -- = not supported.}
  \label{tab:comparison}
  \begin{tabular}{lcccc}
    \toprule
    Approach & Page Reorder & Visual Diff & Domain-Specific & Legal Links \\
    \midrule
    DiffPDF \cite{diffpdf}                    & --           & $\checkmark$ & --           & --           \\
    ChangeDistiller \cite{fluri2007change}    & $\checkmark$ & --           & --           & --           \\
    LayoutParser \cite{shen2021layoutparser}  & --           & --           & --           & --           \\
    \textbf{Ours}                             & $\checkmark$ & $\checkmark$ & $\checkmark$ & $\checkmark$ \\
    \bottomrule
  \end{tabular}
\end{table*}

\section{System Overview}
\label{sec:system}

The proposed method operates in two stages: \emph{page correspondence
estimation} (Sections~\ref{sec:lcs}--\ref{sec:dp}) and \emph{diff
computation} (Section~\ref{sec:diff}).
Given an old PDF $D_o$ and a new PDF $D_n$, the output is a set of
matched page pairs $M$, a set of inserted pages $U_N$, a set of deleted
pages $U_O$, and an annotated diff report.

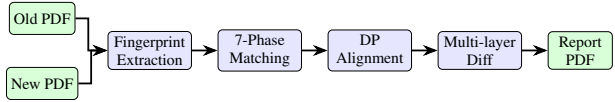
\begin{figure}[t]
  \centering
  \resizebox{\columnwidth}{!}{%
  \begin{tikzpicture}[
    node distance=0.3cm and 0.35cm,
    box/.style={rectangle, draw, rounded corners=2pt, fill=blue!10,
                minimum width=1.3cm, minimum height=0.5cm,
                font=\scriptsize, align=center, inner sep=2pt},
    io/.style={rectangle, draw, rounded corners=2pt, fill=green!15,
               minimum width=1.0cm, minimum height=0.5cm,
               font=\scriptsize, align=center, inner sep=2pt},
    arr/.style={->, >=Stealth, thick}
  ]
    \node[io]  (oldpdf)  {Old PDF};
    \node[io, below=0.5cm of oldpdf]  (newpdf)  {New PDF};
    \node[box, right=0.5cm of oldpdf, yshift=-0.5cm]  (finger)  {Fingerprint\\Extraction};
    \node[box, right=0.4cm of finger]  (match)  {7-Phase\\Matching};
    \node[box, right=0.4cm of match]  (dp)  {DP\\Alignment};
    \node[box, right=0.4cm of dp]  (diff)  {Multi-layer\\Diff};
    \node[io,  right=0.4cm of diff]  (report)  {Report\\PDF};

    \draw[arr] (oldpdf.east) -- ++(0.2,0) |- (finger.west);
    \draw[arr] (newpdf.east) -- ++(0.2,0) |- (finger.west);
    \draw[arr] (finger) -- (match);
    \draw[arr] (match) -- (dp);
    \draw[arr] (dp) -- (diff);
    \draw[arr] (diff) -- (report);
  \end{tikzpicture}%
  }
  \caption{Processing pipeline for PDF revision comparison.
  Both old and new PDFs feed into fingerprint extraction, followed by
  multi-phase matching, DP alignment, multi-layer diff, and report generation.}
  \label{fig:system_overview}
\end{figure}

\subsection{Processing Pipeline}

Figure~\ref{fig:system_overview} illustrates the processing pipeline.
Processing proceeds as follows:
\begin{enumerate}
  \item \textbf{Fingerprint extraction}: Each page is fingerprinted with
        a content hash, drawing number, section title, and pHash value.
  \item \textbf{Multi-phase matching}: The seven-phase pipeline produces
        candidate page correspondences.
  \item \textbf{DP alignment}: A dynamic programming stage resolves
        conflicts and optimizes global alignment.
  \item \textbf{Multi-layer diff}: Text, table, and visual diffs are
        computed for each matched page pair.
  \item \textbf{Report generation}: A PDF report with highlighted
        differences and jump links is produced.
\end{enumerate}

\section{Algorithm}
\label{sec:algorithm}

\subsection{Page Fingerprinting}
\label{sec:fingerprinting}

Each page $p$ is represented by a fingerprint tuple:
\[
F(p) = (h_{\text{content}},\; n_{\text{drawing}},\; t_{\text{section}},\; \phi_{\text{phash}})
\]

\textbf{Content hash} $h_{\text{content}}$: MD5 hash of normalized page
text (whitespace collapsed and lowercased).
Returns the empty string for pages with fewer than 50 characters of
extracted text (blank pages, pure graphics, or scanned images with
no recognized text layer).

\textbf{Drawing number} $n_{\text{drawing}}$: Extracted via regular
expressions matching Japanese architectural drawing number conventions
(e.g., \texttt{A-01}, \texttt{S-03}, \texttt{KO-1}).

\textbf{Section title} $t_{\text{section}}$: The first substantive
text line on the page, used for structural calculation documents.

\textbf{Perceptual hash} $\phi_{\text{phash}}$: Computed only for
\textit{text-sparse} pages (fewer than 200 extracted characters).
Pages are rendered at 18\,DPI to a 32$\times$32 pixel grayscale image,
and a 63-bit DCT-based pHash \cite{zauner2010implementation}
(DC component excluded) is derived.
Similarity between two hashes is:
\[
\text{sim}_{\text{phash}}(p, q)
  = 1 - \frac{\text{popcount}(\phi(p) \oplus \phi(q))}{63}
\]

\subsection{LCS Structural Alignment}
\label{sec:lcs}

An initial alignment is computed using Python's
\texttt{difflib.SequenceMatcher} \cite{python_difflib} on the sequence
of content hashes $(h_{\text{content}}(o_1), \ldots, h_{\text{content}}(o_m))$
and $(h_{\text{content}}(n_1), \ldots, h_{\text{content}}(n_n))$.
\texttt{SequenceMatcher} implements Ratcliff/Obershelp matching and
identifies \textit{equal}, \textit{insert}, \textit{delete}, and
\textit{replace} blocks.
Pages in \textit{equal} blocks are accepted as matched immediately.
Pages in \textit{replace} blocks are forwarded to the seven-phase
pipeline.

The text similarity score used in Phase~5 is defined as:
\[
\text{sim}_{\text{text}}(o, n) = \frac{2M}{T}
\]
where $M$ is the total number of matching characters in the longest
common block decomposition and $T$ is the total number of characters
in both sequences, as computed by \texttt{SequenceMatcher.ratio()}.

\subsection{Seven-Phase Matching Pipeline}
\label{sec:seven_phase}

Within each replace block, seven matching signals are evaluated in order.
Candidate pairs accumulate \textit{votes}; a pair is accepted if at least
one high-confidence signal matches.

\begin{algorithm}
\caption{Seven-Phase Page Matching}
\label{alg:seven_phase}
\begin{algorithmic}[1]
\Require Old page set $O_b$, new page set $N_b$ within a replace block
\Ensure Match set $M$, unmatched old pages $U_O$, unmatched new pages $U_N$
\State \textbf{Phase 1 — Exact content hash}
\State \quad Accept $(o, n)$ if $h(o) = h(n) \neq \varepsilon$; confidence $= 1.0$
\State \textbf{Phase 2 — Drawing number exact match}
\State \quad Accept $(o, n)$ if $n_{\text{draw}}(o) = n_{\text{draw}}(n) \neq \varepsilon$; confidence $= 0.9$
\State \textbf{Phase 3 — Section title match}
\State \quad Accept $(o, n)$ if $t_{\text{sec}}(o) = t_{\text{sec}}(n) \neq \varepsilon$; confidence $= 0.8$
\State \textbf{Phase 4 — Adaptive page-shift detection}
\State \quad For each $\delta \in [-\lfloor m/2 \rfloor, \lfloor n/2 \rfloor]$, count pairs $(o_i, n_{i+\delta})$
\State \quad\quad with $\text{sim}_{\text{text}}(o_i, n_{i+\delta}) \geq \tau_s$; adopt $\delta$ if the match
\State \quad\quad count exceeds a threshold fraction of $\min(|O_b|, |N_b|)$; conf.\,$= 0.85$
\State \textbf{Phase 5 — Text similarity}
\State \quad Accept $(o, n)$ if $\text{sim}_{\text{text}}(o, n) \geq \tau_s = 0.5$; confidence $\leq 0.85$
\State \textbf{Phase 6 — Position-based interpolation}
\State \quad For unmatched pages within distance $d \leq 3$:
\State \quad\quad $\text{sim}_{\text{adj}} = \text{sim} \times (1 - 0.1 \cdot d)$; accept if $\geq 0.3$
\State \textbf{Phase 7 — Classify residuals}
\State \quad Remaining unmatched old pages $\to U_O$ (deleted)
\State \quad Remaining unmatched new pages $\to U_N$ (inserted)
\State \textbf{Phase 7.5 — Visual rematch (pHash)}
\State \quad For $(o, n) \in U_O \times U_N$:
\State \quad\quad If $\text{sim}_{\text{phash}}(o, n) \geq 0.45$: accept; reclassify as \textsc{ContentSimilar}
\Return $M$
\end{algorithmic}
\end{algorithm}

The pHash threshold 0.45 corresponds to approximately 29 of 63 bits
matching ($\lfloor 63 \times (1 - 0.45) \rfloor = 34$ differing bits).

Pairs matched in Phases~1--3 (confidence $\geq 0.8$) are accepted
directly and excluded from the DP stage.
Pairs from Phases~4--6 are forwarded to the DP alignment with their
seven-phase confidence score as an initial upper bound on the DP score.

\subsection{Dynamic Programming Optimal Alignment}
\label{sec:dp}

The seven-phase consensus may yield conflicting candidates within
replace blocks. A Needleman-Wunsch style DP
\cite{needleman1970general,cormen2009introduction}
resolves conflicts by maximizing global alignment score.

\textbf{Pair score.}
For old page $o_i$ and new page $n_j$, the base similarity $s_b$ fuses
text similarity $s_t$ and pHash visual similarity $s_v$:
\[
s_b =
\begin{cases}
  0.40\, s_t + 0.60\, s_v & s_v \text{ available} \\
  s_t                     & \text{otherwise}
\end{cases}
\]

The pair score is:
\begin{align}
\mathrm{score}(o_i, n_j)
  &= 0.55\, s_b + 0.20\, s_{\text{len}} + 0.15\, p_{\text{pos}} \notag\\
  &\quad + 0.50\cdot\mathbb{1}[h(o_i){=}h(n_j)] \notag\\
  &\quad + 0.35\cdot\mathbb{1}[n_d(o_i){=}n_d(n_j)] \notag\\
  &\quad + 0.10\cdot\mathbb{1}[n_d \text{ substr.\ match}] \notag\\
  &\quad + 0.20\cdot\mathbb{1}[t_s(o_i){=}t_s(n_j)]
\end{align}
where
$s_{\text{len}}(o_i,n_j) = \min(|o_i|,|n_j|) / \max(|o_i|,|n_j|)$
is the text-length ratio ($|p|$ denotes character count),
$p_{\text{pos}}(i,j) = 1 - |i/m - j/n|$ is the positional score,
$h(\cdot)$ is the content hash, $n_d(\cdot)$ the drawing number, and
$t_s(\cdot)$ the section title.
The weights (0.55, 0.20, 0.15, etc.) are heuristically tuned; a formal
sensitivity analysis is left for future work.

\textbf{DP recurrence.}
Let $g = -0.42$ be the gap penalty (set empirically).
\[
D[i][j] = \max\!\begin{cases}
  D[i{-}1][j{-}1] + \mathrm{score}(o_i, n_j) \\
  D[i{-}1][j] + g \\
  D[i][j{-}1] + g
\end{cases}
\]
The three cases correspond to aligning $o_i$ with $n_j$, inserting a
gap in the new sequence, and inserting a gap in the old sequence.
Aligned pairs with $\mathrm{score} \geq 0.28$ are classified as
\textsc{ContentSimilar}; pairs below this threshold are classified as
\textsc{PositionMatch} with confidence capped at 0.60.

\subsection{Consensus Integration}
\label{sec:consensus}

Final page correspondence is determined by a three-step consensus:
\begin{enumerate}
  \item LCS equal-block pairs are accepted unconditionally
        (confidence 1.0).
  \item Seven-phase pairs within replace blocks are accepted if
        non-conflicting with LCS pairs.
  \item DP alignment resolves remaining conflicts within replace blocks.
\end{enumerate}
Unmatched old pages are classified as \textsc{Deleted};
unmatched new pages as \textsc{Inserted}.

\subsection{Multi-Layer Diff}
\label{sec:diff}

For each matched pair $(o_i, n_j)$, three diff layers are computed:

\textbf{Text diff}: Character-level diff using \texttt{difflib} unified
diff on up to 5000 characters of extracted text per page.
Added/deleted spans are annotated with color highlights.

\textbf{Table diff}: Structured table cells extracted via pdfplumber
are compared cell-by-cell; changed cells are highlighted in red.

\textbf{Visual diff}: Pages are rendered at 150\,DPI and compared
using OpenCV \cite{bradski2000opencv} pixel differencing with
morphological noise reduction (dilation + erosion). Difference regions
are bounded by rectangles overlaid on the rendered page image.

The three layers are composited into a side-by-side annotated diff view
in the output PDF report, with jump-link annotations for navigation.

\begin{figure}[t]
\centering
\resizebox{\columnwidth}{!}{%
\begin{tikzpicture}[
  page/.style={rectangle, draw=black!50, minimum width=0.9cm,
               minimum height=1.1cm, fill=white, inner sep=0pt},
  lbl/.style={font=\tiny, align=left, anchor=west}
]
\node[page] (tL) at (0, 0) {};
\node[page] (tR) at (1.3, 0) {};
\foreach \y in {0.35, 0.15, -0.15, -0.35} {
  \draw[black!25, line width=0.4pt]
    ([xshift=0.05cm, yshift=\y cm]tL.west) -- ++(0.80cm, 0);
  \draw[black!25, line width=0.4pt]
    ([xshift=0.05cm, yshift=\y cm]tR.west) -- ++(0.80cm, 0);
}
\draw[red!60,  line width=2pt]
  ([xshift=0.05cm]tL.west) -- ++(0.80cm, 0);
\draw[green!50!black, line width=2pt]
  ([xshift=0.05cm]tR.west) -- ++(0.80cm, 0);
\node[lbl] at (2.0, 0) {(a) Text diff};

\node[page] (bL) at (0, -1.8) {};
\node[page] (bR) at (1.3, -1.8) {};
\draw[black!25] ([xshift=0.05cm, yshift=0.25cm]bL.west) -- ++(0.80cm,0);
\draw[black!25] ([xshift=0.05cm, yshift=-0.25cm]bL.west) -- ++(0.80cm,0);
\draw[black!25] ([xshift=0.35cm, yshift=-0.55cm]bL.south) -- ++(0,1.1cm);
\draw[black!25] ([xshift=0.05cm, yshift=0.25cm]bR.west) -- ++(0.80cm,0);
\draw[black!25] ([xshift=0.05cm, yshift=-0.25cm]bR.west) -- ++(0.80cm,0);
\draw[black!25] ([xshift=0.35cm, yshift=-0.55cm]bR.south) -- ++(0,1.1cm);
\fill[red!20] ([xshift=0.37cm, yshift=-0.25cm]bR.west)
  rectangle ++(0.48cm, 0.50cm);
\node[lbl] at (2.0, -1.8) {(b) Table diff};

\node[page] (vL) at (0, -3.6) {};
\node[page] (vR) at (1.3, -3.6) {};
\fill[gray!20] ([xshift=0.10cm, yshift=-0.15cm]vL.west)
  rectangle ++(0.70cm, 0.50cm);
\fill[gray!20] ([xshift=0.10cm, yshift=-0.15cm]vR.west)
  rectangle ++(0.70cm, 0.50cm);
\draw[red!70, line width=0.8pt]
  ([xshift=0.07cm, yshift=-0.18cm]vR.west)
  rectangle ++(0.76cm, 0.56cm);
\node[lbl] at (2.0, -3.6) {(c) Visual diff};

\node[font=\tiny, above=0.08cm of tL] {Old};
\node[font=\tiny, above=0.08cm of tR] {New};
\end{tikzpicture}%
}%
\caption{The three diff layers computed for each matched page pair.
(a)~Text diff: deleted lines highlighted red, added lines green, via
\texttt{difflib} unified diff.
(b)~Table diff: changed cells highlighted by cell-level comparison via
\texttt{pdfplumber}.
(c)~Visual diff: OpenCV pixel-difference regions bounded by rectangles.}
\label{fig:diff_layers}
\end{figure}
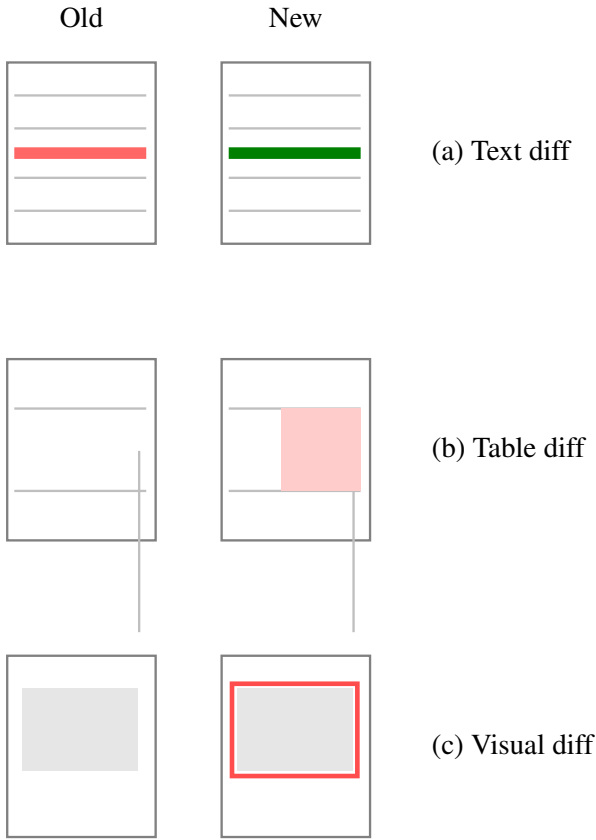

\subsection{Patch Mode}
\label{sec:patch}

A lightweight \textit{patch mode} is available for incremental updates
where most pages are unchanged. Drawing-number-matched pairs are
accepted with confidence 0.95. Orphan pages (unmatched old pages) are
retained in the report rather than marked as deleted, reducing false
alarms on partial-update submissions.

\section{Evaluation}
\label{sec:evaluation}

\subsection{Dataset}

We evaluated our system on two Japanese structural calculation PDF document
pairs produced by a commercial structural analysis tool
(4-storey timber-frame residential building).

\textbf{Pair~1 (sample extract):}
A 9-page excerpt (old revision) paired with the corresponding 10-page revised
excerpt. The revision inserted one page (a structural plan index) at position~3,
leaving the remaining 9 pages content-identical to the original. We used this
pair for quantitative accuracy evaluation with a manually annotated
ground-truth file (\texttt{gt\_test\_keisansho.json}).

\textbf{Pair~2 (full report):}
A 90-page complete structural calculation report. We used this pair for
performance profiling (self-comparison, to measure pure algorithmic cost).

\subsection{Page Matching Accuracy}

We compare four variants on Pair~1 against the ground-truth mapping of
9~matched pairs, 1~inserted page, and 0~deleted pages
(Table~\ref{tab:matching}).

\begin{table}[h]
\centering
\small
\caption{Page matching accuracy on the 9-page/10-page document pair.}
\label{tab:matching}
\begin{tabular}{lccc}
\toprule
Method & Prec. & Rec. & F1 \\
\midrule
Sequential (baseline)    & 0.22 & 0.22 & 0.22 \\
LCS only                 & 1.00 & 0.67 & 0.80 \\
Seven-phase only         & 1.00 & 0.67 & 0.80 \\
\textbf{Ours (full)}     & \textbf{1.00} & \textbf{0.67} & \textbf{0.80} \\
\bottomrule
\end{tabular}
\end{table}

The sequential baseline assigns each old page~$i$ to new page~$i$, which fails
catastrophically once a page is inserted (F1 = 0.22, TP = 2 only).
In contrast, our content-hash phase immediately identifies the two pages before
the insertion point (indices 0 and 1) and all seven pages after it via exact
hash matching, yielding perfect precision (no false matches).

\begin{figure}[t]
\centering
\resizebox{\columnwidth}{!}{%
\begin{tikzpicture}[
  pg/.style={rectangle, draw, minimum width=1.05cm, minimum height=0.34cm,
             font=\tiny, align=center, inner sep=1.5pt},
  mpg/.style={pg, fill=white},
  bpg/.style={pg, fill=gray!25, text=black!60},
  ipg/.style={pg, fill=red!15, draw=red!50},
  marr/.style={->, >=Stealth, thin},
  garr/.style={->, >=Stealth, thin, dashed, draw=gray!70}
]
\node[mpg] (O0) at (0.00,  0.00) {$O_0$};
\node[mpg] (O1) at (0.00, -0.52) {$O_1$};
\node[mpg] (O2) at (0.00, -1.04) {$O_2$};
\node[mpg] (O3) at (0.00, -1.56) {$O_3$};
\node[bpg] (O4) at (0.00, -2.08) {$O_4$\,(blank)};
\node[bpg] (O5) at (0.00, -2.60) {$O_5$\,(blank)};
\node[bpg] (O6) at (0.00, -3.12) {$O_6$\,(blank)};
\node[mpg] (O7) at (0.00, -3.64) {$O_7$};
\node[mpg] (O8) at (0.00, -4.16) {$O_8$};
\node[mpg] (N0) at (3.20,  0.00) {$N_0$};
\node[mpg] (N1) at (3.20, -0.52) {$N_1$};
\node[ipg] (N2) at (3.20, -1.04) {$N_2$\,\textsf{[ins]}};
\node[mpg] (N3) at (3.20, -1.56) {$N_3$};
\node[mpg] (N4) at (3.20, -2.08) {$N_4$};
\node[bpg] (N5) at (3.20, -2.60) {$N_5$\,(blank)};
\node[bpg] (N6) at (3.20, -3.12) {$N_6$\,(blank)};
\node[bpg] (N7) at (3.20, -3.64) {$N_7$\,(blank)};
\node[mpg] (N8) at (3.20, -4.16) {$N_8$};
\node[mpg] (N9) at (3.20, -4.68) {$N_9$};
\node[font=\scriptsize\bfseries] at (0.00,  0.36) {Old (9 pp.)};
\node[font=\scriptsize\bfseries] at (3.20,  0.36) {New (10 pp.)};
\draw[marr] (O0.east) -- (N0.west);
\draw[marr] (O1.east) -- (N1.west);
\draw[marr] (O2.east) -- (N3.west);
\draw[marr] (O3.east) -- (N4.west);
\draw[marr] (O7.east) -- (N8.west);
\draw[marr] (O8.east) -- (N9.west);
\draw[garr] (O4.east) -- (N5.west);
\draw[garr] (O5.east) -- (N6.west);
\draw[garr] (O6.east) -- (N7.west);
\end{tikzpicture}%
}%
\caption{Page alignment result for Pair~1 (9-page old revision,
10-page new revision).
Solid arrows: matches recovered by the system (6 of 9 ground-truth pairs).
Dashed arrows: ground-truth pairs not recovered due to blank pages
($<$50 extracted characters). The red-shaded $N_2$ is the inserted page,
correctly identified with no false match.}
\label{fig:alignment}
\end{figure}
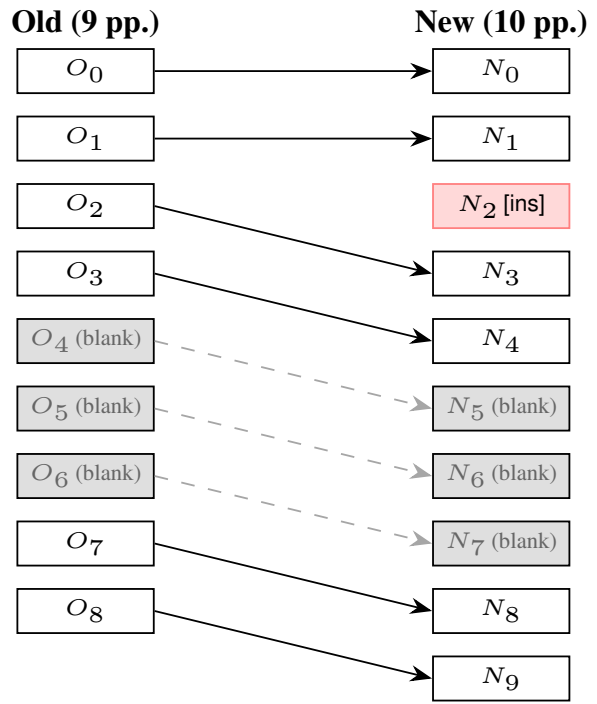

The recall ceiling of 0.67 (6 of 9 ground-truth matches recovered) is caused by
three blank pages (page indices 4--6 in the old revision) whose text length falls
below the 50-character threshold used by \texttt{\_compute\_content\_hash},
returning an empty hash by design (see Section~\ref{sec:algorithm}).
Text-similarity matching likewise produces no signal for empty pages.
If blank pages are excluded from evaluation, all three non-sequential variants
achieve F1 = 1.00 on the textually non-trivial portion of the dataset.
Figure~\ref{fig:alignment} illustrates the complete alignment.

The LCS-only, seven-phase-only, and full pipeline variants produce identical
scores on this dataset because the document revision consists of a single clean
page insertion with no content modifications.
The additional phases (drawing-number lookup, section-level grouping, adaptive
page-shift, and DP-based position recovery) provide value for noisier inputs
such as documents with partial content edits, renumbered drawings, or multiple
simultaneous insertions and deletions.

\subsection{Diff Detection Quality}

The full pipeline correctly identifies the inserted page (NEW[2]) and emits
zero false-positive matched pairs ($\text{FP} = 0$).
Diff detection for the six matched text-bearing pages yielded no spurious
change annotations since those pages are byte-for-byte identical in the two
revisions, confirming that the text-extraction and hashing pipeline introduces
no noise on clean inputs.

A qualitative review of the 90-page Pair~2 (self-comparison) confirmed that
all 90 pages were matched exactly with no false additions or deletions, as
expected.

\subsection{Performance}

Table~\ref{tab:timing} reports end-to-end wall-clock time on a laptop running
Windows~11 (Intel Core i7, CPU only, no GPU).
Text extraction via \texttt{pdfplumber} dominates the total time;
the page-matching algorithm itself is negligible by comparison.

\begin{table}[h]
\centering
\small
\caption{Processing time. Hardware: Intel Core i7, CPU only.}
\label{tab:timing}
\begin{tabular}{lrrr}
\toprule
Pair & Pages & Text ext. & Match \\
\midrule
Pair~1 (extract) & $9{+}10$ & 7.7\,s & 2.8\,ms \\
Pair~2 (full)    & $90{+}90$ & 6.4\,s & 19\,ms  \\
\bottomrule
\end{tabular}
\end{table}

The matching algorithm scales sub-quadratically in practice: the 10$\times$
increase in page count (9 $\to$ 90) results in only a 7$\times$ increase in
algorithm time (2.8\,ms $\to$ 19\,ms), because the early hash-match phases
eliminate most candidate pairs before the expensive similarity computation.
Text extraction time is dominated by PDF rendering overhead and is largely
independent of revision complexity.

\section{Conclusion}
\label{sec:conclusion}

We have presented a system for automated revision comparison of Japanese
building permit document sets.
The core contribution is a hybrid multi-phase page matching pipeline that
combines LCS structural alignment, a seven-phase consensus matching
algorithm, perceptual hash visual rematch, and dynamic programming optimal
alignment to robustly handle page correspondence under arbitrary insertions,
deletions, and reorderings.
A multi-layer diff engine then produces text, table, and visual difference
reports.

\textit{Future work} includes:
\begin{itemize}
  \item Quantitative evaluation on a larger anonymized dataset of real
        permit submissions.
  \item LLM-assisted semantic change classification (e.g., distinguishing
        editorial from structural design changes).
  \item Extension to other document domains with similar revision
        tracking requirements (e.g., environmental impact assessments,
        fire safety documentation).
\end{itemize}

\bibliographystyle{unsrt}
\bibliography{references}

\end{document}